\documentclass[10pt,twocolumn,letterpaper]{article}

\usepackage[pagenumbers]{cvpr}

\usepackage[T1]{fontenc}
\usepackage[utf8]{inputenc}
\usepackage{microtype}
\usepackage{amsfonts,bm,mathtools}
\usepackage{multirow,array,makecell}
\usepackage{tikz}
\usetikzlibrary{arrows.meta,positioning,fit,backgrounds,calc,shapes.geometric,decorations.pathreplacing}

\definecolor{tzblue}{HTML}{1F4E79}
\definecolor{tzaccent}{HTML}{C25E00}
\definecolor{tzgrey}{HTML}{6B7280}
\definecolor{tzlight}{HTML}{EEF2F7}
\definecolor{tzexpert}{HTML}{2E6F40}
\definecolor{tznull}{HTML}{8B1A1A}

\usepackage[breaklinks=true,colorlinks=true,linkcolor=tzblue,citecolor=tzblue,urlcolor=tzblue]{hyperref}

\newcommand{\softmax}{\operatorname{softmax}}
\newcommand{\topk}{\operatorname{Top}\text{-}k}
\newcommand{\clip}{\operatorname{clip}}
\newcommand{\SiLU}{\operatorname{SiLU}}

\newcommand{\Nexp}{N}
\newcommand{\model}{Wonder\,4}

\title{\model{}: A Mixture-of-Experts Recipe for Image Quality Enhancement}

\author{Yunan Wu\quad Bin Sun\quad Partha Acharjee\quad Xiaoyu Wang\\
Topaz Labs Research\\
}

\hypersetup{pdftitle={Wonder 4},pdfauthor={Yunan Wu, Bin Sun, Partha Acharjee, Xiaoyu Wang}}

\begin{document}
\maketitle

\begin{abstract}
A real photograph is often blurry, grainy, compressed, scratched, and low-resolution at the same time. Each defect has its own severity and its own place in the image. The corrections for these defects conflict with each other. Existing models describe the whole image with one description, so they cannot say where each correction belongs.
To solve this problem, we present \textbf{\model{}}, a mixture-of-experts (MoE) model for image quality enhancement. It corrects several defects of a real photograph at the same time, each in the region where it occurs. Every block of the model holds a mixture of $\Nexp = 13$ experts.
A router, trained with the model, sends every image patch to $k = 5$ of them. No degradation classifier, mask, prompt, detector, or task label is used.
Another issue is that standard MoE training, taken from language models, pushes every expert toward equal use. Equal use applies every correction everywhere, even where it is not needed. We replace this rule with a bias that only sets a minimum share of use. Across all patches, an expert used less than this minimum gets higher router scores until it is selected again. An expert above the minimum is not changed, so no expert is forced to be used as often as the others. The bias needs no extra loss. The method adds no extra module and no extra processing step.
We evaluate \model{} on DIV2K and Flickr2K, against five published methods and three open-weight pipelines. It reaches the best LPIPS, DISTS, and NIQE on all three test settings (DIV2K $\times 4$, DIV2K $\times 8$, and Flickr2K $\times 4$).
The full version of the report will be released soon.
\end{abstract}

\newcommand{\teaserpass}{enhanced result}
\newcommand{\teasercap}{in one pass}
\newcommand{\teaserlq}{LQ image}
\providecommand{\teaserpass}{one pass}
\providecommand{\teasercap}{in a single step}
\providecommand{\teaserlq}{LQ image}
\begin{figure}[t]
\centering
\resizebox{\linewidth}{!}{
\begin{tikzpicture}[
  x=1cm,y=1cm,
  tile/.style={inner sep=0pt,outer sep=0pt},
  lab/.style={font=\scriptsize,text=black!80,align=center},
  tag/.style={font=\scriptsize\bfseries,text=white,fill=tzblue,rounded corners=2pt,inner sep=2.5pt},
  box/.style={draw=tzgrey,rounded corners=2pt,fill=tzlight,font=\scriptsize,align=center,inner sep=2.5pt},
  arr/.style={->,>=Stealth,thick,tzblue!85},
  cell/.style={draw=tzgrey!70,minimum size=2.6mm,inner sep=0pt,outer sep=0pt},
  on/.style={cell,fill=tzexpert},
  off/.style={cell,fill=white},
  tok/.style={draw=tzblue!60,fill=tzblue!12,minimum size=1.9mm,inner sep=0pt,outer sep=0pt}
]
\node[tile] (rt)  at (2.75,0)  {\includegraphics[width=3.05cm]{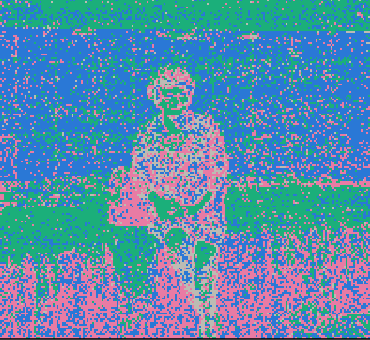}};
\node[tile] (in)  at (-0.05,0) {\includegraphics[width=2.35cm]{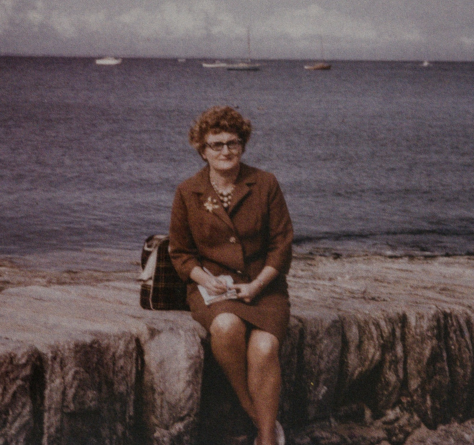}};
\node[tile] (out) at (5.55,0)  {\includegraphics[width=2.35cm]{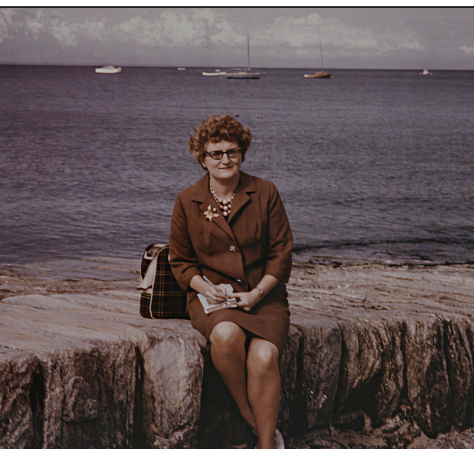}};
\node[lab,anchor=north] at (-0.05,-1.45) {\textbf{input}\\faded, soft, low-res};
\node[lab,anchor=north] at (2.75,-1.45)  {\textbf{per-patch routing} (one block)\\color = first-preference expert};
\node[lab,anchor=north] at (5.55,-1.45)  {\textbf{\model{}}\\\teaserpass{}};
\draw[arr] (in.east) -- (rt.west);
\draw[arr] (rt.east) -- (out.west);
\node[box,minimum height=8mm,minimum width=14mm] (lq) at (-0.55,-3.38) {\teaserlq{}};
\node[draw=tzblue!60,rounded corners=3pt,fill=white,minimum width=43mm,minimum height=20mm] (blk) at (2.75,-3.38) {};
\node[font=\scriptsize,text=tzblue,anchor=north] at ([yshift=-1pt]blk.north) {network block $\times L$};
\foreach \r in {0,1,2}{\foreach \c in {0,1,2}{
  \node[tok] at ({1.05+\c*0.21},{-3.05-\r*0.21}) {};}}
\node[box,minimum width=11mm,minimum height=3.4mm,fill=tzaccent!15,draw=tzaccent] (router) at (2.75,-3.26) {router};
\draw[arr,thin] (1.62,-3.26) -- (router.west);
\foreach \i/\s in {1/off,2/on,3/off,4/off,5/on,6/off,7/on,8/off,9/off,10/on,11/off,12/on,13/off}{
  \node[\s] (e\i) at ({2.75-2.10+\i*0.30},-3.80) {};
}
\draw[arr,thin] (router.south) -- (2.75,-3.63);
\node[font=\tiny,text=black!70,anchor=north] at (2.75,-3.97) {13 experts, 5 active per patch};
\node[box,minimum height=8mm,minimum width=14mm] (hq) at (6.05,-3.38) {enhanced\\image};
\draw[arr] (lq.east) -- (blk.west);
\draw[arr] (blk.east) -- (hq.west);
\node[tag] at (-0.55,-4.15) {no task label};
\node[tag] at (6.05,-4.15) {no detector};
\end{tikzpicture}}
\caption{\textbf{\model{} at a glance.} \emph{Top:} a low-quality (LQ) input photograph, a map showing which expert
each patch (a small region of the image) prefers in one block, and the enhanced output. Colors are expert indices.
\emph{Bottom:} the LQ image passes through $L$ blocks. In each block a router sends every patch to 5 of 13 experts (green cells).
Because every patch chooses its own experts, a photograph with several defects is corrected \teasercap{} with no
task label and no degradation detector.}
\label{fig:teaser}
\end{figure}
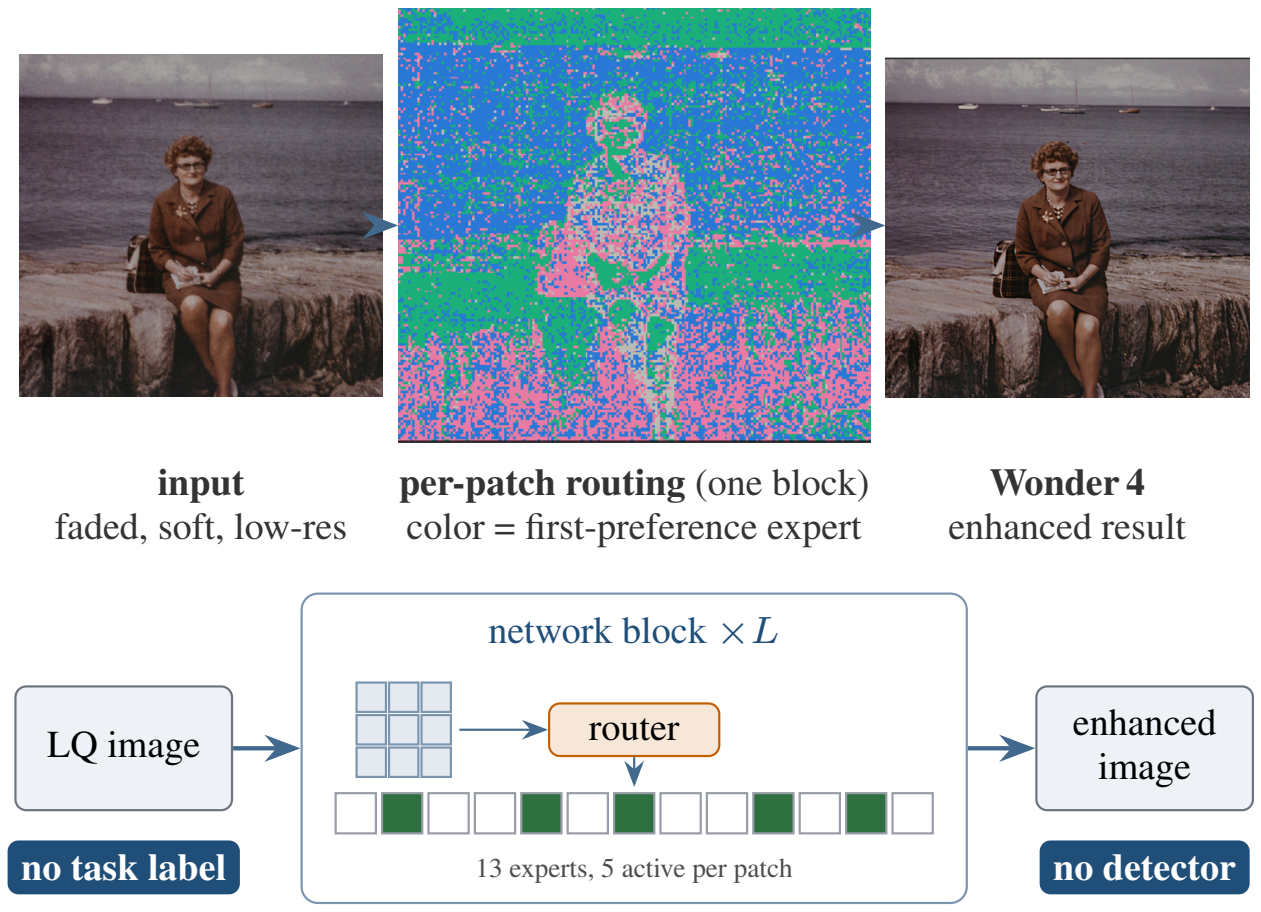

\section{Introduction}
\label{sec:intro}

A scanned family photograph is usually blurry from the lens and the scan, grainy from the film,
marked by compression bands, covered by dust and scratches from the print, and much lower in
resolution than people now want to view. Every defect is present at the same time. Each defect
has its own severity and is limited to its own part of the image. This is the ordinary case in
consumer photography, and we call it \emph{high-order mixed degradation}. The required output
quality is much higher. A result is acceptable only if it still looks correct at full resolution,
with no softened edge, no invented texture, and no visible boundary between corrected and
untouched regions. That is the quality a photo retoucher aims for, and the quality we mean by
\emph{professional-grade}. These corrections conflict with each other. Removing noise also removes
film grain. Keeping the grain leaves the blur uncorrected. Removing a scratch adds false texture
if the same correction is applied to a clean sky. A model that improves this photograph must decide
not only how strongly to retouch each region but \emph{where}, and it must apply incompatible
corrections to different regions of one image.

Task-specific models for super-resolution, denoising, deblurring, compression-artifact
removal and damage repair are each trained on a degradation pipeline built for
one task and have never seen this combination of defects. Running them one after another has three
drawbacks: the order of the corrections must be chosen in advance, each model works on the invented
detail of the one before it, and a detector is needed to decide which model runs. Unified models were introduced to solve this problem, but they all stop for the same reason. They condition on a \emph{global} description of the
degradation, whether inferred as a code~\citep{li2022airnet} or
prompt~\citep{potlapalli2023promptir}, borrowed from a vision--language
model~\citep{luo2024daclip}, written or planned as text~\citep{conde2024instructir,liu2025fapeir},
predicted as the predominant class~\citep{jiang2024autodir}, or set by a task
flag~\citep{lin2024diffbir}. One
description for the whole image cannot say which region each correction belongs to, and getting that description
needs a second network, whose mistakes limit the whole result. This limitation also shows
in how the field evaluates these models. The standard test combines denoising, deraining,
dehazing, deblurring and low-light enhancement~\citep{zhang2025aioirsurvey}, and the
corrections missing from that test are exactly the \emph{localized} ones.
Neither pipelines of separate models nor these unified models handle high-order mixed degradation. A pipeline never sees the combined
defects and cannot apply conflicting corrections without detectors and a fixed order.
A unified model that conditions on one global description cannot assign incompatible
corrections to different regions of the picture.

Because the image can be split into patches, a mixture of experts (MoE) is a good match: it computes a
\emph{combination} of corrections at every patch, rather than one decision about which task
is present. In language modeling the standard design is already fixed. An MoE network replaces each
dense layer with an MoE layer of structurally identical experts and a linear
router that selects a few experts for each input~\citep{shazeer2017moe,fedus2022switch,jiang2024mixtral,deepseekv3,yang2025qwen3,openai2025gptoss},
and the same layer is used without change in generative image
models~\citep{fei2024ditmoe}. Image restoration has not adopted it. Existing restoration
mixtures either select experts once per image or replace the experts with
prompts, adapters, or whole models chosen by an external degradation
estimator~\citep{zamfir2025moceir,kong2026mimdit,ren2024moediffir,he2025mor,wang2024unirestorer,li2025m2restore}.
Most of them also take a rule from language models: every expert should be used equally. This is
right when the experts do the same job, and harmful when a correction belongs on only a few patches.
Scratch removal is one such case, since a scratch covers only a small part of the image.

We propose \model{}, which is the first MoE architecture for image quality enhancement.
Its experts are $\Nexp = 13$ structurally identical small networks, one MoE layer per block
of the backbone, selected $k = 5$ at a time per patch by a
router local to that block. The router needs no estimator, so each patch is routed
where its own defect occurs. One rule adapts the recipe to photographs. We add a
bias to the scores the router uses when it chooses experts~\citep{wang2024auxfree,deepseekv3}.
Each expert must be selected for at least a minimum share of the patches. If an expert is selected
less often than this minimum, that bias increases until the router selects it often enough. Once an expert is above the
minimum, the bias stops increasing, so a correction that belongs in only a few places stays
there.
Other models cover many defects by adding extra parts. \model{} uses the standard mixture and adds nothing else: no second network to classify the damage,
no prompts, no mask, no second processing step, and no separate path for each task. \model{} is released in a consumer application.

\newcommand{\wfourbank}[3]{
  \foreach \i in {1,...,13}{
    \pgfmathtruncatemacro{\bcol}{mod(\i-1,5)}
    \pgfmathtruncatemacro{\brow}{(\i-1)/5}
    \pgfmathsetmacro{\bx}{#1+\bcol*0.205}
    \pgfmathsetmacro{\by}{#2-\brow*0.205}
    \draw[fill=tzgrey!12,draw=tzgrey!45,line width=0.25pt]
      (\bx,\by) rectangle ++(0.15,0.15);
  }
  \foreach \i in {#3}{
    \pgfmathtruncatemacro{\bcol}{mod(\i-1,5)}
    \pgfmathtruncatemacro{\brow}{(\i-1)/5}
    \pgfmathsetmacro{\bx}{#1+\bcol*0.205}
    \pgfmathsetmacro{\by}{#2-\brow*0.205}
    \draw[fill=tzexpert!75,draw=tzexpert,line width=0.25pt]
      (\bx,\by) rectangle ++(0.15,0.15);
  }
}

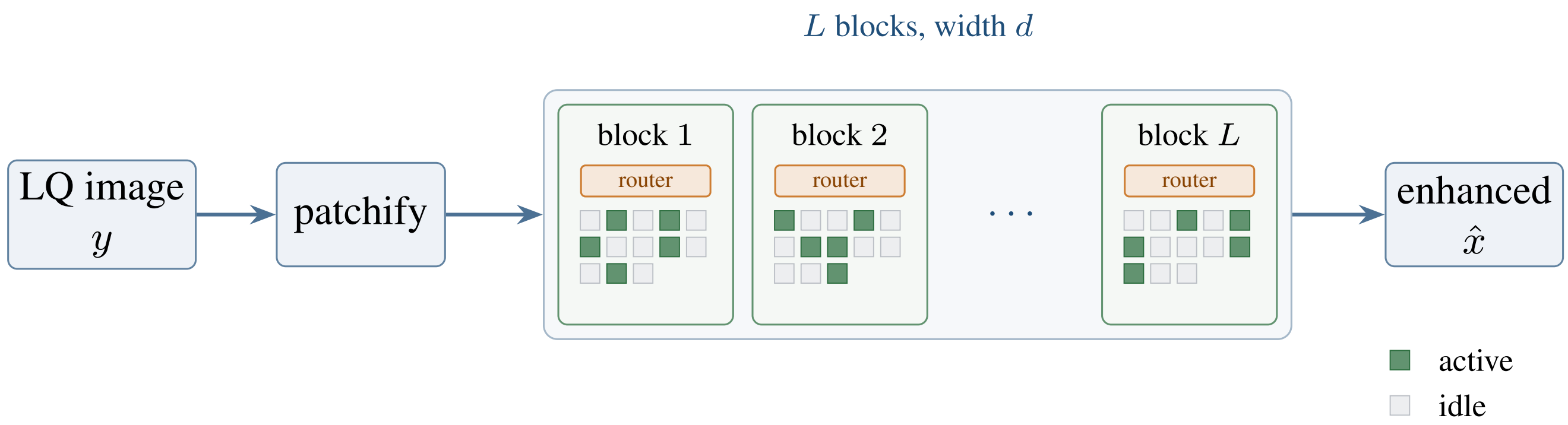
\begin{figure*}[t]
\centering
\resizebox{\linewidth}{!}{
\begin{tikzpicture}[
  font=\small, x=1cm, y=1cm,
  >={Stealth[length=2mm]},
  box/.style={draw=tzblue!70,fill=tzlight,rounded corners=2pt,align=center,inner sep=2.5pt},
  blk/.style={draw=tzexpert!75,fill=tzexpert!5,rounded corners=2pt},
  rtr/.style={draw=tzaccent!80,fill=tzaccent!14,rounded corners=1pt,
              font=\tiny,inner sep=1pt,text=tzaccent!70!black},
  flow/.style={->,thick,tzblue!80},
  side/.style={->,thick,tzaccent,dashed}
]

\node[box,minimum width=13mm,minimum height=8mm] (lq)  at (0.00,0) {LQ image\\$y$};
\node[box,minimum width=13mm,minimum height=8mm] (patch) at (2.00,0) {patchify};

\node[blk,minimum width=13.5mm,minimum height=17mm] (b1) at (4.20,0) {};
\node[font=\scriptsize] at (4.20,0.62) {block $1$};
\node[rtr,minimum width=10mm,minimum height=2.4mm] at (4.20,0.26) {router};
\wfourbank{3.695}{-0.12}{2,4,6,9,12}

\node[blk,minimum width=13.5mm,minimum height=17mm] (b2) at (5.70,0) {};
\node[font=\scriptsize] at (5.70,0.62) {block $2$};
\node[rtr,minimum width=10mm,minimum height=2.4mm] at (5.70,0.26) {router};
\wfourbank{5.195}{-0.12}{1,4,7,8,13}

\node[blk,minimum width=13.5mm,minimum height=17mm] (bL) at (8.40,0) {};
\node[font=\scriptsize] at (8.40,0.62) {block $L$};
\node[rtr,minimum width=10mm,minimum height=2.4mm] at (8.40,0.26) {router};
\wfourbank{7.895}{-0.12}{3,5,6,10,11}

\node[tzblue] at (7.05,0) {$\cdots$};
\begin{scope}[on background layer]
  \node[draw=tzblue!40,fill=tzblue!4,rounded corners=3pt,
        fit=(b1)(bL),inner sep=3pt] (ditbox) {};
\end{scope}
\node[font=\scriptsize,tzblue,anchor=south] at (6.30,1.22) {$L$ blocks, width $d$};

\node[box,minimum width=13mm,minimum height=8mm] (out) at (10.60,0) {enhanced\\$\hat x$};

\draw[flow] (lq) -- (patch);
\draw[flow] (patch) -- (ditbox);
\draw[flow] (ditbox) -- (out);

\draw[fill=tzexpert!75,draw=tzexpert,line width=0.25pt] (9.95,-1.20) rectangle ++(0.15,0.15);
\node[font=\scriptsize,anchor=west] at (10.20,-1.125) {active};
\draw[fill=tzgrey!12,draw=tzgrey!45,line width=0.25pt] (9.95,-1.55) rectangle ++(0.15,0.15);
\node[font=\scriptsize,anchor=west] at (10.20,-1.475) {idle};

\end{tikzpicture}}
\caption{\textbf{\model{} inference path.} The low-quality (LQ) input image $y$ is split into patches (\emph{patchify}) and
passed through a backbone of $L$ blocks of width $d$, which produces the enhanced image $\hat x$. Every block carries its own router and its own bank of $N{=}13$
experts (small cells). For each patch the router activates $k{=}5$ of them (green, \emph{active}); the rest stay idle (grey). The active cells drawn differ between blocks
because selection is made per patch and per block, not once per image.}

\label{fig:arch}
\end{figure*}

\section{Model}
\label{sec:method}

A photograph that needs several corrections, each in its own region, cannot be handled by one decision about the whole
image. It needs a decision at every location, and a way to apply more than one correction there. A sparse mixture
of experts does both, so \model{} uses its standard form. An MoE layer of identical experts replaces the standard dense layer of each block. A linear
router is placed in each layer. It selects the top $k$ experts for each patch. To achieve better performance, we change one thing in \model{}: each expert is
given a minimum share of use instead of an equal share, so a correction is applied only where it is needed (\cref{sec:optimization}). This section first describes the architecture,
then the MoE layer, the router, expert training, and the training objective.

\subsection{Model Architecture}
\label{sec:backbone}

\Cref{fig:arch} shows \model{}. It takes the low-quality input image $y$, resized to the target resolution, and
produces the enhanced image $\hat x$. A backbone network with $L$ blocks of width $d$ does the work. It operates on
patches of the image. An
MoE layer with its own router replaces the standard dense layer in every block.
This backbone fits a per-patch mixture. Splitting the image into patches gives every region its
own unit, so each region can be routed separately. No second network is needed to estimate the damage.

\subsection{Mixture-of-Experts Layer}
\label{sec:architecture}

In every block $\ell$, an MoE layer of $\Nexp = 13$ experts replaces the ordinary dense layer. Each expert is a small network of intermediate width $d_{\mathrm{ff}}$, in the gated form called SwiGLU,
\begin{equation}
E^{(\ell)}_a(u) \;=\; W^{(\ell)}_{a,3}\Big(\SiLU\big(W^{(\ell)}_{a,1}\, u\big)\odot W^{(\ell)}_{a,2}\, u\Big),
\label{eq:expert_ffn}
\end{equation}
Here $a$ indexes the expert, $\ell$ the block, $u$ is the input of the layer at one patch and the $W$'s are learned matrices. $\SiLU$ is a smooth on--off function, and $\odot$ multiplies two vectors entry by entry.
All experts are built the same way and start from random weights. Each patch uses $k = 5$ of the 13. In each layer the router selects a new set of 5 experts.
This lets the model hold more experts without using more computation at run time. Each expert is specialized for a specific type of defect. Five experts per patch let corrections
combine where defects overlap.

\subsection{Router and Top-\texorpdfstring{$k$}{k} Selection}
\label{sec:routing}

The router decides where each correction belongs. Each block has its own router, and the router works on one patch at a time. For patch $i$ it applies one linear layer to the layer's input $u^{(\ell)}_i$,
\begin{equation}
o^{(\ell)}_i = W^{(\ell)}_r\, u^{(\ell)}_i + b^{(\ell)}_r,
\qquad g^{(\ell)}_i = \softmax\big(o^{(\ell)}_i\big).
\label{eq:router}
\end{equation}
Here $i$ indexes the patch, and $\softmax$ turns the scores $o$ into probabilities $g$ over the experts. The routing decision is made by this one linear layer.
The layer is not trained to classify the type of damage.

The choice of which experts a patch uses is kept separate from how much each selected expert contributes. The extra score
$\beta^{(\ell)}_a$, a per-expert bias that raises the selection chance of under-used experts (\cref{sec:optimization}), is added only to the scores used to choose experts. It is not added to the
probabilities. In training, random noise is also added to those scores. The noise lets the router try other experts~\citep{shazeer2017moe},
\begin{equation}
\mathcal{A}^{(\ell)}_i = \topk_{a}\big(o^{(\ell)}_{i,a} + \beta^{(\ell)}_a + \sigma_{\mathrm{ex}}\,\xi_{i,a}\big),
\quad \xi_{i,a}\sim\mathcal{N}(0,1),
\label{eq:topk}
\end{equation}
Here $\mathcal{A}^{(\ell)}_i$ is the set of selected experts, $\topk$ keeps the $k$ largest scores, and $\xi$ is random noise of scale $\sigma_{\mathrm{ex}}$, with $\sigma_{\mathrm{ex}} = 0$ at inference. The probabilities of the selected experts are scaled so they sum
to one~\citep{jiang2024mixtral}. The layer returns their weighted sum in place of the dense layer's output $\tilde u$,
\begin{equation}
\tilde u^{(\ell)}_i = \sum_{a\in\mathcal{A}^{(\ell)}_i} w^{(\ell)}_{i,a}\, E^{(\ell)}_a\!\big(u^{(\ell)}_i\big),
\quad w^{(\ell)}_{i,a} = \frac{g^{(\ell)}_{i,a}}{\sum_{a'\in\mathcal{A}^{(\ell)}_i} g^{(\ell)}_{i,a'}}.
\label{eq:blockout}
\end{equation}
The weights always sum to one. The extra score can change which experts a patch uses, and the amount of correction stays the same. \Cref{eq:expert_ffn,eq:router,eq:topk,eq:blockout} are the full computation of the
MoE layer.

\subsection{Expert Training}
\label{sec:optimization}

\begin{figure*}[t]
\centering
\includegraphics[width=\linewidth]{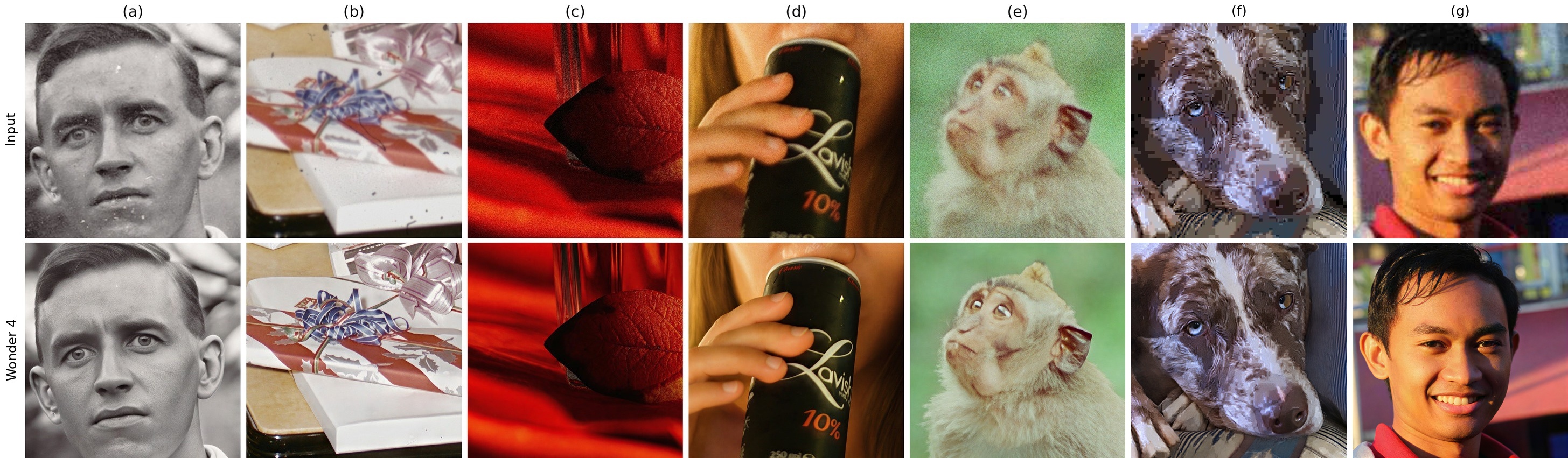}
\caption{\textbf{Showcase: several defects mixed in real photographs.} Each column shows the same
crop from the input (top) and \model{} (bottom). (a)~Scanned print with dust and scratches over a blurred
region. (b)~Dust and scratches on wrapping paper. (c)~Heavy grain in a low-light photograph.
(d)~Grain and softness on fine detail and lettering. (e)~Grain on fur. (f)~Compression blocking and softness on fur.
(g)~Blur and grain together. Every column carries at least two defect families at once, and \model{}
corrects them in a single pass with no detector, mask or task label. Best viewed at full size.}
\label{fig:mixed-real}
\end{figure*}

The usual rule pushes every expert toward equal use~\citep{fedus2022switch}. That rule is right when experts are meant to do the same job, but it is wrong in our case. Some experts are needed in only a small part of the images.
 Under an equal-use rule, an expert that is needed on only a small subset of patches, such as the physical-damage expert, is also applied to patches where it should not be used. Stars in a night sky are one such case. The same bright points are physical damage anywhere else in a photograph, and a user expects them to be removed, but in a photograph of the night sky they are the subject and must not be erased.
 Every expert then learns the same average correction. We keep the usual balancing mechanism, a bias on the selection scores, and change its
target. Each expert has a minimum share of the routing weight, summed over all patches. The router still selects only $k = 5$ experts for each patch. An expert below its minimum gets higher selection scores until it is selected often enough, so no expert is left unused.
 An expert above its minimum is not changed, so no expert is forced to be used more often than the training images need.

The mechanism is a bias from DeepSeek-V3~\citep{wang2024auxfree,deepseekv3}. It needs no extra loss. It acts only
on the scores used for selection, so this balancing step does not work against the main training loss. The load of expert $a$ in
block $\ell$ is the share of the routing weight it receives over the $T$ patches of one training step,
$f^{(\ell)}_a = \frac1T\sum_{i} w^{(\ell)}_{i,a}$. We track it with an exponential moving average $\bar f^{(\ell)}_a$, a running average that gives more weight to recent steps, with decay $\gamma$.
The floor $\phi$ is a fraction $\rho$ of an equal share, $\rho/\Nexp$.
The bias grows with step size $\eta_\beta$ and stops growing once an expert reaches the floor ($\clip$ limits the ratio to $[0,1]$),
\begin{equation}
\beta^{(\ell)}_a \leftarrow \beta^{(\ell)}_a + \eta_\beta\,
\clip\!\left(\frac{\phi - \bar f^{(\ell)}_a}{\phi},\,0,\,1\right).
\label{eq:floor}
\end{equation}
After each update, the biases of a block are shifted to average zero and limited to $[-\kappa,\kappa]$. An expert
below the floor is increased until that expert is selected often enough to be trained. An expert above the floor is not changed.

\subsection{Training Objective}
\label{sec:objective}

The model is trained end to end in a single run. Three small loss terms are added to the enhancement loss. The ST-MoE
router $z$-loss keeps the router's scores from growing too large~\citep{zoph2022stmoe}. An entropy penalty on the
routing weights, increased gradually during training, pushes each patch to give most of its weight to only a few of its five experts. A third
penalty applies when the cosine similarity of two experts' weights exceeds a threshold $c_0$. This penalty keeps experts from
learning the same correction. Experts can still overlap where their corrections should overlap. The
enhancement loss is a weighted sum of mean squared error (MSE) terms, LPIPS~\citep{zhang2018lpips} and
an adversarial term, which rewards realistic images.

\section{Experiments}
\label{sec:experiments}

The experiments compare \model{} with existing methods on public benchmarks, test it on real photographs
that carry several defects at once, and measure whether the router makes a separate decision at each location.

\definecolor{mapblue}{RGB}{42,120,214}
\definecolor{mappurple}{RGB}{74,58,167}
\definecolor{mapgreen}{RGB}{27,175,122}
\definecolor{mappink}{RGB}{232,123,164}
\definecolor{mapgrey}{RGB}{188,187,179}
\begin{figure*}[t]
\centering
\resizebox{\textwidth}{!}{%
\begin{tikzpicture}[
  x=1cm,y=1cm,
  tile/.style={inner sep=0pt,outer sep=0pt},
  lab/.style={font=\scriptsize,anchor=south,inner sep=1pt,align=center},
  rowtag/.style={font=\small\bfseries,anchor=east},
  sw/.style={draw=black!40,minimum size=2.4mm,inner sep=0pt,outer sep=0pt},
  lt/.style={font=\scriptsize,anchor=west,inner sep=1pt,xshift=2pt}
]
\def\H{2.6}
\node[rowtag] at (-0.15,\H/2) {(a)};
\node[tile,anchor=south west] (ain) at (0,0) {\includegraphics[height=\H cm]{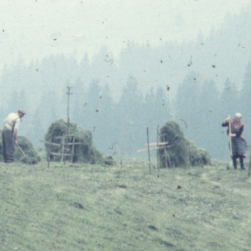}};
\node[lab] at (ain.north) {input (square crop)\\$1\times$ upscale, $130\times130$ patches};
\node[tile,anchor=south west] (a0)  at ({\H+0.15},0)                 {\includegraphics[height=\H cm]{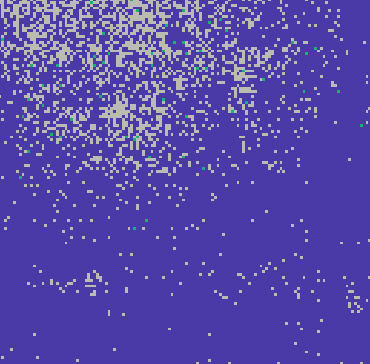}};
\node[tile,anchor=south west] (a10) at ({\H+0.15+1*(\H+0.15)},0)       {\includegraphics[height=\H cm]{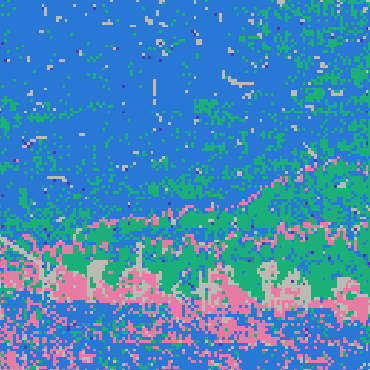}};
\node[tile,anchor=south west] (a15) at ({\H+0.15+2*(\H+0.15)},0)       {\includegraphics[height=\H cm]{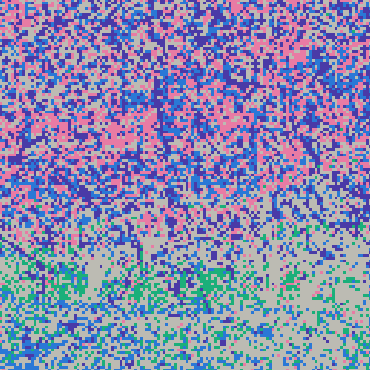}};
\node[tile,anchor=south west] (a24) at ({\H+0.15+3*(\H+0.15)},0)       {\includegraphics[height=\H cm]{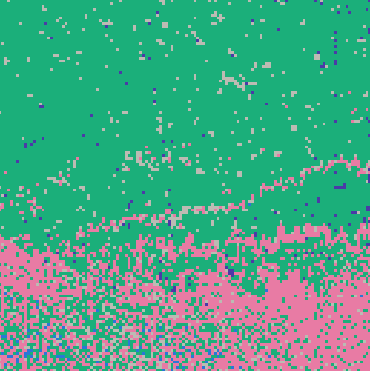}};
\node[tile,anchor=south west] (a29) at ({\H+0.15+4*(\H+0.15)},0)       {\includegraphics[height=\H cm]{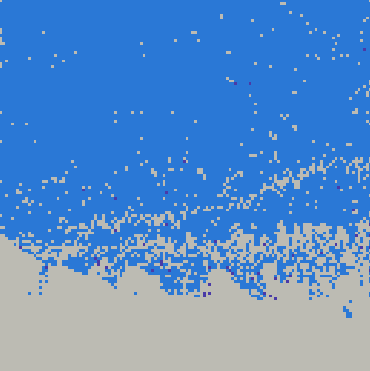}};
\foreach \node/\label in {a0/0,a10/10,a15/15,a24/24,a29/29}{
  \node[lab] at (\node.north) {block \label};
}
\begin{scope}[shift={({\H+0.15},-0.22)}]
  \foreach \c/\t [count=\i from 0] in {mapblue/expert 10,mappurple/expert 7,mapgreen/expert 8,mappink/expert 5,mapgrey/other experts}{
    \node[sw,fill=\c,anchor=north west] (s\i) at ({\i*1.85},0) {};
    \node[lt] at (s\i.east) {\t};
  }
\end{scope}
\def\Y{-\H-1.15}
\node[rowtag] at (-0.15,{\Y+\H/2}) {(b)};
\node[tile,anchor=south west] (bin) at (0,\Y) {\includegraphics[height=\H cm]{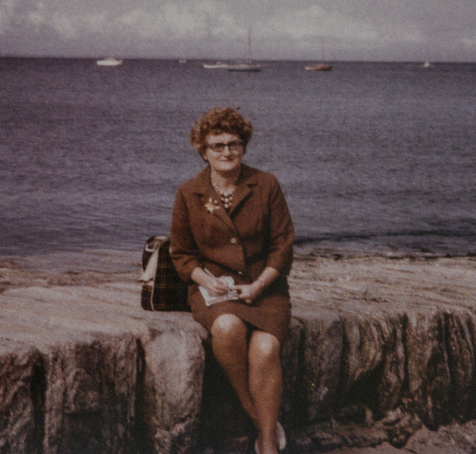}};
\node[lab] at (bin.north) {input\\$2\times$ upscale, $190\times199$ patches};
\node[tile,anchor=south west] (b0)  at ({2.72+0.15},\Y)                 {\includegraphics[height=\H cm]{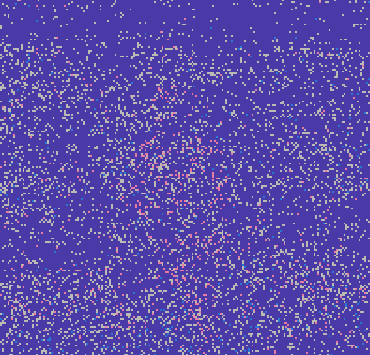}};
\node[tile,anchor=south west] (b2)  at ({2.72+0.15+1*(\H+0.15)},\Y)       {\includegraphics[height=\H cm]{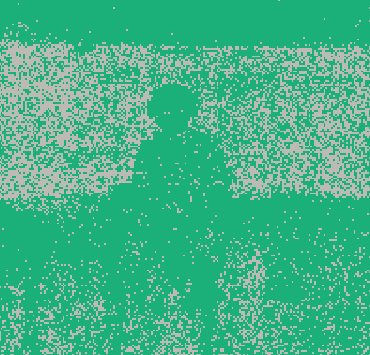}};
\node[tile,anchor=south west] (b10) at ({2.72+0.15+2*(\H+0.15)},\Y)       {\includegraphics[height=\H cm]{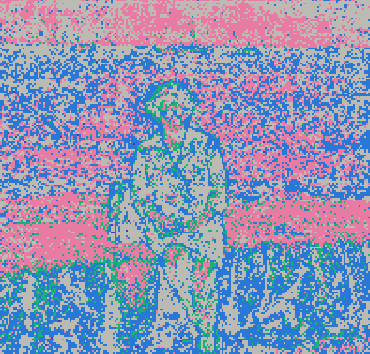}};
\node[tile,anchor=south west] (b15) at ({2.72+0.15+3*(\H+0.15)},\Y)       {\includegraphics[height=\H cm]{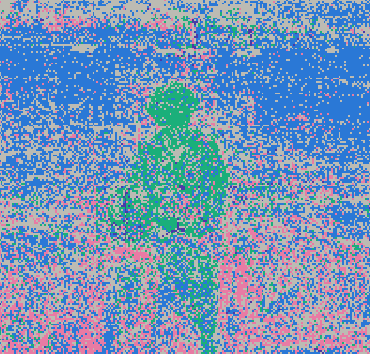}};
\node[tile,anchor=south west] (b28) at ({2.72+0.15+4*(\H+0.15)},\Y)       {\includegraphics[height=\H cm]{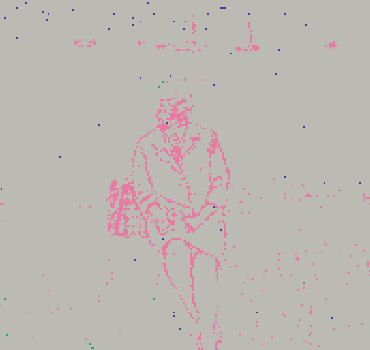}};
\foreach \node/\label in {b0/0,b2/2,b10/10,b15/15,b28/28}{
  \node[lab] at (\node.north) {block \label};
}
\begin{scope}[shift={(2.87,{\Y-0.22})}]
  \foreach \c/\t [count=\i from 0] in {mapblue/expert 13,mappurple/expert 7,mapgreen/expert 9,mappink/expert 10,mapgrey/other experts}{
    \node[sw,fill=\c,anchor=north west] (t\i) at ({\i*1.85},0) {};
    \node[lt] at (t\i.east) {\t};
  }
\end{scope}
\end{tikzpicture}}
\caption{\textbf{Per-patch routing, drawn on the image.} Rows (a) and (b) are two photographs; the first tile of each
row is the input (a is a square center crop; the grid size is given in patches, and the upscale factor is 1$\times$ for (a) and 2$\times$ for (b)).
Each other tile is a map of one block, numbered from $0$ at the network input. A patch is colored by the expert the
router ranks first for it, aligned with the input. Early blocks are flat; late blocks follow the scene patch by patch.
Each row has its own color key: only its four most-used experts are colored, all others are grey, and experts are
identified only by their index.}
\label{fig:maps}
\end{figure*}
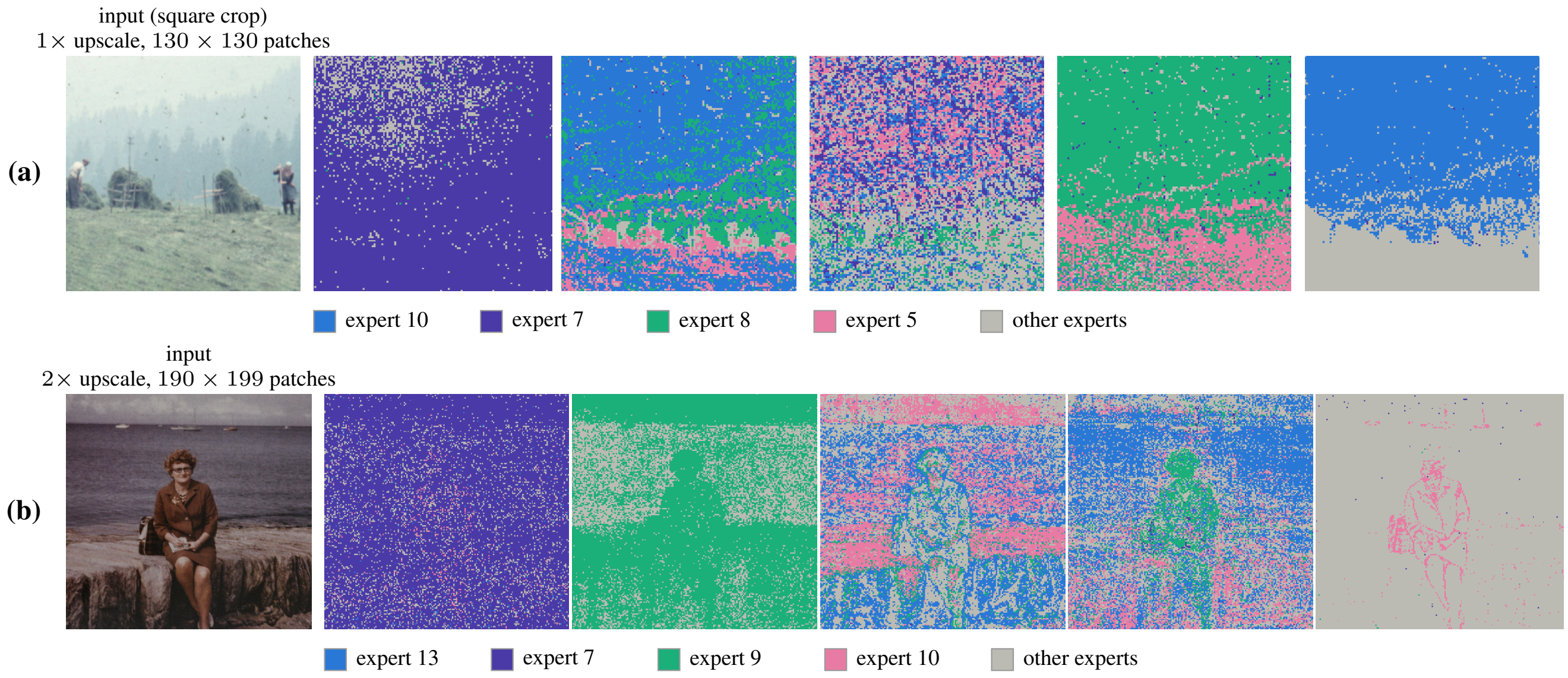

\subsection{Experimental Settings}

\textbf{Benchmarks.} We use three test settings. The first is DIV2K validation ($100$ images) under the mixed
$\times 4$ degradation of the NTIRE 2018 challenge, called wild $\times 4$~\citep{agustsson2017div2k,timofte2018ntire}. Blur, noise and
compression are mixed, and their strength varies from image to image. The second is DIV2K validation at
$\times 8$ bicubic downsampling ($100$ images). At this scale, simply resizing the image is not enough, so methods must
create the missing detail. The third is Flickr2K at $\times 4$ ($2\,650$ images)~\citep{lim2017edsr}. This set is
large enough that a difference in the average score is reliable. No DIV2K or Flickr2K image was used to train \model{}.
Several baselines, for example Real-ESRGAN, include both datasets in their training data.

\textbf{Metrics.} We report PSNR on the Y (brightness) channel, where higher is better, and the perceptual metrics LPIPS~\citep{zhang2018lpips},
DISTS~\citep{ding2020dists} and NIQE~\citep{mittal2013niqe}. LPIPS and DISTS compare with the ground truth, the original sharp image. NIQE does not.

\textbf{Compared methods.} We compare against five published enhancement systems, Real-ESRGAN~\citep{wang2021realesrgan},
StableSR~\citep{wang2024stablesr} (the authors' few-step Turbo variant), SeeSR~\citep{wu2024seesr},
SUPIR~\citep{yu2024supir} and OSEDiff~\citep{wu2024osediff}, and three widely adopted open-weight pipelines, the
FLUX.1 ControlNet upscaler~\citep{labs2024flux}, the Stable Diffusion $\times 4$ upscaler~\citep{rombach2022ldm}
and Qwen-Image-Edit~\citep{wu2025qwenimage}. All baselines run at their authors' default settings.

\subsection{Quantitative Results}

\Cref{tab:main} gives the comparison. \model{} reaches the best LPIPS, DISTS
and NIQE on every test setting, against both the published methods and the open-weight pipelines. The gains are largest on Flickr2K. There it lowers LPIPS by
$19\%$ and DISTS by $27\%$ relative to the strongest published baseline. Among the open-weight pipelines,
Qwen-Image-Edit is the closest on the metrics that use the ground truth. \model{} still improves on it by $18\%$
in LPIPS and $16\%$ in DISTS on DIV2K $\times 4$, by $25\%$ and $22\%$ on DIV2K $\times 8$, and by $10\%$ and $15\%$
on Flickr2K.

\begin{table*}[t]
\centering
\footnotesize
\setlength{\tabcolsep}{3.2pt}
\caption{\textbf{Quantitative comparison} on DIV2K validation (100 images, $\times 4$ wild and $\times 8$ bicubic) and Flickr2K ($2\,650$ images, $\times 4$ unknown degradation). Rows: five published enhancement methods (top), three open-weight pipelines (middle) and \model{} (bottom). Columns: PSNR (dB, on the brightness channel), LPIPS and DISTS (both compare with the ground-truth image) and NIQE (needs no ground truth). $\uparrow$/$\downarrow$: higher/lower is better. Best per column over all methods in \textbf{bold}, second best \underline{underlined}. The full nine-metric tables are in the full report.}
\label{tab:main}
\resizebox{\linewidth}{!}{
\begin{tabular}{l rrrr rrrr rrrr}
\toprule
& \multicolumn{4}{c}{DIV2K wild $\times 4$} & \multicolumn{4}{c}{DIV2K bicubic $\times 8$} & \multicolumn{4}{c}{Flickr2K unknown $\times 4$} \\
\cmidrule(lr){2-5}\cmidrule(lr){6-9}\cmidrule(lr){10-13}
Method & PSNR$\uparrow$ & LPIPS$\downarrow$ & DISTS$\downarrow$ & NIQE$\downarrow$ & PSNR$\uparrow$ & LPIPS$\downarrow$ & DISTS$\downarrow$ & NIQE$\downarrow$ & PSNR$\uparrow$ & LPIPS$\downarrow$ & DISTS$\downarrow$ & NIQE$\downarrow$ \\
\midrule
Real-ESRGAN~\citep{wang2021realesrgan} & 18.96 & 0.4047 & 0.2107 & 4.36 & 23.43 & 0.3998 & 0.2016 & 5.80 & \underline{22.07} & 0.3723 & 0.1887 & 4.46 \\
StableSR (Turbo)~\citep{wang2024stablesr} & \textbf{19.09} & 0.4424 & 0.2186 & 6.05 & \textbf{24.09} & 0.4013 & 0.1998 & 5.78 & 22.04 & 0.3922 & 0.1950 & 5.83 \\
SeeSR~\citep{wu2024seesr} & 18.91 & 0.3641 & \underline{0.1612} & 3.75 & 23.46 & 0.2927 & \underline{0.1341} & 3.66 & 22.02 & 0.3387 & 0.1649 & 3.76 \\
SUPIR~\citep{yu2024supir} & \underline{19.07} & 0.3698 & 0.1849 & 3.82 & 23.42 & 0.3014 & 0.1490 & 3.66 & \textbf{22.64} & 0.3965 & 0.2194 & 4.75 \\
OSEDiff~\citep{wu2024osediff} & 18.67 & \underline{0.3479} & 0.1629 & \underline{3.59} & 23.10 & \underline{0.2842} & 0.1355 & \underline{3.61} & 21.24 & 0.3391 & 0.1618 & 4.02 \\
\midrule
FLUX.1 ControlNet Upscaler~\citep{labs2024flux} & 19.04 & 0.4094 & 0.1908 & 4.19 & 22.40 & 0.3382 & 0.1530 & 3.95 & 21.01 & 0.4152 & 0.1949 & 4.12 \\
SD $\times 4$ Upscaler~\citep{rombach2022ldm} & 18.07 & 0.6153 & 0.2892 & 3.94 & 22.49 & 0.4482 & 0.1977 & 3.85 & 21.66 & 0.4302 & 0.1914 & \underline{3.68} \\
Qwen-Image-Edit~\citep{wu2025qwenimage} & 18.97 & 0.3991 & 0.1684 & 4.17 & 22.66 & 0.3092 & 0.1385 & 4.03 & 21.96 & \underline{0.3061} & \underline{0.1381} & 4.15 \\
\midrule
\textbf{\model{}} & 18.69 & \textbf{0.3282} & \textbf{0.1419} & \textbf{3.53} & \underline{23.62} & \textbf{0.2318} & \textbf{0.1075} & \textbf{2.93} & 20.41 & \textbf{0.2745} & \textbf{0.1177} & \textbf{3.24} \\
\bottomrule
\end{tabular}}
\end{table*}

\model{} does not have the highest PSNR. The methods with the highest PSNR change the input the least, and they rank in the lower half on
the perceptual metrics. This is the usual trade-off. A model that adds detail looks better and gets a worse
PSNR. SeeSR and OSEDiff, the closest published competitors, lead the learned scores that do not use the ground
truth (MUSIQ, CLIP-IQA, MANIQA and TOPIQ) on DIV2K $\times 4$ in the full tables. They add more fine detail than
the ground-truth image contains. Those scores reward the extra detail, and metrics that compare with the ground truth give it
a lower score.

\subsection{Real Photographs with Mixed Defects}

\Cref{fig:mixed-real} shows real photographs with several defects at once: dust and scratches on scanned prints,
heavy grain from low light, compression blocks on fur, and blur together with grain. The test settings above
are super-resolution settings. They test only one kind of correction. Real photographs test the mixture. Here, \model{} applies the
corrections each region needs, together with upscaling, in one pass. It uses no detector, mask or task label.

\subsection{Localized Expert Routing}
\label{sec:analysis}

A mixture only helps if the router makes a separate decision at each location. A router that selects the same experts everywhere,
or one set of experts for the whole image, could still get a good score while failing to do this. We inspect the routing directly. Each block
makes one routing decision per patch, so the routing of a block can be drawn as one map. For every patch we take
the expert with the highest routing probability, its first choice, and place it on the patch grid, lined up with
the photograph (\Cref{fig:maps}). Experts are identified by index only. We assign no job to any of them.

The maps become more detailed in later blocks. The earliest blocks use almost the same expert everywhere and ignore the image content. In \cref{fig:maps}(a), block~$0$ makes
expert~$7$ the first choice for $87\%$ of patches. Structure that follows the image appears soon after. At
block~$2$ of (b), expert~$9$ takes $85\%$ of the patches on the person in photo (b) but only $67\%$ of the sea and sky. The
person forms one region, and the background is a mix of experts. In the later blocks the split follows the scene, patch by
patch. In (a), blocks~$24$ and~$29$ both separate the meadow from the line of trees. The two blocks use different pairs of experts,
and they keep the boundary within three patch rows of the same contour. In (b), at block~$28$,
expert~$10$ follows the person's outline, one patch wide. It takes $18\%$ of patches on the person and under $1\%$
in the sky. The router is never given a segmentation. Every boundary in the map is the router's own estimate of where the
image changes.

The same effect can be measured without the picture. Writing $w_i$ for the routing weights of
\cref{eq:blockout} and $\bar w = \frac1T\sum_i w_i$ for their average over the $T$ patches of one image, the
within-image routing mutual information ($H$ is entropy)
\begin{equation}
I^{(\ell)} \;=\; H\big(\bar w^{(\ell)}\big) - \tfrac{1}{T}\textstyle\sum_i H\big(w^{(\ell)}_i\big)
\;\in\; \big[0,\ \log\Nexp\big]
\label{eq:mi}
\end{equation}
measures how much a patch's routing depends on which patch it is, inside one image. Averaged over all $L$
blocks, $I \approx 0.89$ nats (natural-log units), against a maximum of $\log 13 \approx 2.56$. Being above zero matters as much as the exact value, because a value of zero would mean there is no per-patch decision.
A router that ignores location, and one that decides once per image, both give $I \approx 0$. A value clearly
above zero can only come from patches of the same image being routed differently. A correction can change from
place to place only when that is true.

\section{Conclusion}
\label{sec:conclusion}

A real photograph has several defects at the same time, and the corrections conflict. A single model for every defect was limited by
two things. One description per image cannot say where a correction belongs. Equal use of experts applies every correction everywhere, even where it is not needed.
\model{} removes both limits. It keeps the standard sparse mixture: $\Nexp = 13$ experts per block, and
$k = 5$ selected per patch by a router. Each expert
gets a minimum share of use instead of equal use. No classifier, estimator, mask, or task label is used.
On DIV2K and Flickr2K, \model{} reaches the best LPIPS, DISTS, and NIQE on every test setting, against published methods
and open-weight pipelines. The routing maps (\cref{fig:maps}) show decisions made inside each image. The method does not depend
on our network, our data, or our set of defects. We offer it as a simple, effective baseline for sparse mixtures
in image enhancement.

\bibliographystyle{plainnat}

\end{document}